\documentclass{article}

\usepackage{PRIMEarxiv}

\usepackage[utf8]{inputenc} 
\usepackage[T1]{fontenc}    
\usepackage{hyperref}       
\usepackage{url}            
\usepackage{booktabs}       
\usepackage{amsfonts}       
\usepackage{nicefrac}       
\usepackage{microtype}      
\usepackage{lipsum}
\usepackage{fancyhdr}       
\usepackage{graphicx}       
\usepackage[dvipsnames]{xcolor}
\usepackage{csquotes}
\usepackage{color,soul}
\usepackage{wrapfig}
\usepackage{caption}
\usepackage{subcaption}
\usepackage{algorithm,algpseudocode}
\usepackage{amssymb}
\usepackage{graphicx}

\usepackage{blindtext}
\usepackage{array}
\usepackage{chngcntr}
\usepackage{makecell}

\graphicspath{{media/}}     
\usepackage[acronym, nohypertypes={acronym}, nomain]{glossaries}
\newacronym{ml}{ML}{Machine Learning}
\newacronym[first={Text zu Bild (engl. Text-to-Image, txt2img)}]{txt2img}{Txt2Img}{Text-to-Image}

\newacronym{rnn}{RNN}{Recurrent Neural Networks}
\newacronym{vae}{VAE}{Variational Autoencoders}
\newacronym{gan}{GAN}{Generative Adversarial Networks}
\newacronym{hvae}{HVAE}{Hierarchical Variational Autoencoders}
\newacronym{mhvae}{MHVAE}{Markovian \acrshort{hvae}}
\newacronym{snr}{SNR}{Signal-Rausch-Verhältnis}
\newacronym{elbo}{ELBO}{Evidence Lower Bound}
\newacronym{vlb}{VLB}{Variational Lower Bound}
\newacronym{smiles}{SMILES}{Simplified Molecular-Input Line-Entry System}
\newacronym{nlp}{NLP}{Natural Language Processing}
\newacronym{gpt}{GPT}{General Pretrained Transformer}
\newacronym{selfies}{SELFIES}{Self-referencing Embedded String}
\newacronym{lora}{LoRA}{Low-Rank Adaptation}
\newacronym{sascore}{SAScore}{Synthetic Accessibility Score}
\newacronym{ffn}{FFN}{Feed Forward Network}
\newacronym{gqa}{GQA}{Grouped-Query Attention}

\title{LLamol: A Dynamic Multi-Conditional Generative Transformer for De Novo Molecular Design 
}

\author{
  Niklas Dobberstein, Astrid Maaß, Jan Hamaekers \\
  Fraunhofer SCAI \\
  Department of Virtual Material Design (VMD) \\
  Bonn, Germany\\
  \texttt{\{niklas.dobberstein, astrid.maass, jan.hamaekers\}@scai.fraunhofer.de} \\
}

\begin{document}
\maketitle

\begin{abstract}

Generative models have demonstrated substantial promise in \acrfull{nlp} and have found application in designing molecules, as seen in \acrfull{gpt} models. In our efforts to develop such a tool for exploring the organic chemical space in search of potentially electro-active compounds, we present \textit{LLamol}, a single novel generative transformer model based on the LLama 2 architecture, which was trained on a 13M superset of organic compounds drawn from diverse public sources. To allow for a maximum flexibility in usage and robustness in view of potentially incomplete data, we introduce \textit{Stochastic Context Learning} as a new training procedure. We demonstrate that the resulting model adeptly handles single- and multi-conditional organic molecule generation with up to four conditions, yet more are possible. The model generates valid molecular structures in \acrshort{smiles} notation 
while 
flexibly incorporating three numerical and/or one token sequence into the generative process, just as requested. The generated compounds are very satisfactory in all scenarios tested. In detail, we showcase the model's capability to utilize token sequences for conditioning, either individually or in combination with numerical properties, making \textit{LLamol} a potent tool for de novo molecule design, easily expandable with new properties.

\end{abstract}


\section{Introduction}



In fields like energy storage materials or medicinal chemistry, substances are key to technological advancement and progress: the success of these applications hinges on the specific properties of the materials. 
However, the processes of discovery and development of new materials often face practical and/or principal obstacles, such as unavailability of compounds or precursors, high production costs, and the need for extensive trials on the practical side, or limited data and/or experience, as well as biased expectations of designers and developers 
on the other hand.
Generative models, a powerful category in machine learning, have the potential to address both of these issues simultaneously, as they can help focus our efforts a priori only on the \emph{most likely} candidates.

Many architectures related to creation of novel data points 
were developed in recent years, most notably \acrfull{rnn} \cite{rnn}, \acrfull{gan} \cite{gan}, \acrfull{vae} \cite{vae} and Transformers \cite{vaswani2017attention}.
The transformer architecture, especially, has revolutionized the fields of Natural Language Processing (\acrshort{nlp}) \cite{brown2020language} and other domains like computer vision \cite{vit}. 
The introduction of the \acrfull{gpt} architecture led to significant advancements in generative natural language applications.
Generative models have also been applied in the fields of medicine and material science to create new molecules with \emph{predefined} features, a process known as conditional generation \cite{urbina2022megasyn, molgpt}. This application can significantly accelerate the discovery of new candidate molecules.
Although current generative models may not provide the optimal solution, they can greatly reduce the size of the chemical space that needs to be evaluated.
Current estimates for the size of the chemical space containing drug-like molecules range from $10^{23}$ to $10^{60}$ \cite{polishchuk2013estimation}.
Many approaches have successfully used \acrshort{vae}s \cite{richards2022conditional, Lim2018}, \acrshort{gan}s \cite{decao2022molgan}, or \acrshort{rnn}s \cite{Grisoni2020}. However, more recently, transformer models, specifically the GPT models \cite{molgpt, Chen2023}, have emerged as the new state-of-the-art in this domain, especially, in the field of conditional molecular generation 
\cite{Wang2021MultiContraint, Wang2023CMolGPT}. 
A good summary of available models can be found in the survey from Du et. al. \cite{du2022molgensurvey}.

Bagal et al.\cite{molgpt} presented the MolGPT architecture from which a family of models, each one tailored to a specific task, could be derived. Inspired by their work, we set out to develop a \emph{solitary} model that can handle many tasks simultaneously to support 
the search for low-cost, high-energy-density alternatives for energy storage materials in flow batteries. 
The model itself should not require complex training data; thus, it operates on \acrshort{smiles}\cite{smiles_paper} -- a minimalist molecular representation that allows us to draw a mass of data from numerous sources -- and easy to provide and directly to verify target properties that serve as conditions (primarily to facilitate the development process of the model). 
\footnote{A condition, here, is a desired molecular property that we want to provide to the model. Based on this condition, the model should generate new molecules that satisfy the requested value.}



In this paper, we present a new, dynamic training approach termed "Stochastic Context Learning" (SCL) to train a single model for conditional generation, capable of generating molecules as \acrshort{smiles} while respecting a variable number of conditions. 
Our training dataset consists of approx. 13 million organic molecules, which is a 
superset of several public datasets 
(see Section \ref{sec:dataset}). On this, we train a GPT-style transformer model, specifically a model based on LLama 2 \cite{touvron2023llama}, to generate new compounds based on one or more conditions/target property.
To achieve this, 
we assign a learnable embedding to each property value. This ensures that the model perceives not only the numerical value, but also the associated label. 

To be able to assess the model's performance 
directly, we 
chose three easily determined numerical properties: 
SAScore \cite{sascore} (reflecting production cost), and logP and molecular weight (contributing to energy density), along with another optional condition: a user-defined core structure that has to be integrated into the final molecule.  The latter is given as a \acrshort{smiles} string, which is a continuous sequence of tokens, hereafter referred to as a 'token sequence'. 
 \footnote{A token sequence can represent either a complete molecule or a molecular fragment, which may not necessarily be valid independently. However, a token sequence should become part of a valid molecule when incorporated into the generative process. }

In the following sections, we detail the architecture, training data and process along with the results 
obtained for unconditional, single, and multi-conditional molecule generation.



\section{Architecture} \label{sec:architecture}

The architecture we utilized, as depicted in Figure \ref{fig:current_architecture}, is a modified version of the LLama2 architecture \cite{touvron2023llama} 
as obtained from GitHub (https://github.com/karpathy/llama2.c). 
The hyperparameters can be found in Table \ref{table:hyperparams_arch}, which we determined from previous experiments.

Our model consists of approximately 15 million parameters and is composed of eight decoder blocks. Each decoder block includes a masked multi-head self-attention layer, followed by a \acrfull{ffn} 
that employs the SwiGLU \cite{shazeer2020glu} activation function. While the original LLama 2 architecture utilized \acrfull{gqa} 
\cite{ainslie2023gqa}, we opted for the full multi-head attention mechanism given the comparatively smaller size of our model.

The masked multi-head self-attention layer \cite{vaswani2017attention}, defined by Equation \ref{eq:mmha}, takes an embedded input sequence $X \in \mathbb{R}^{L \times d_{emb}}$ of length $L$, where each element represents an embedding vector with dimension $d_{emb}$. 
Through the attention mechanism, each head learns to attend to a different part of the sequence, resulting in an attention matrix $\text{head}_i \in \mathbb{R}^{L \times d_v}$. 
We utilize dot-product self-attention, which produces three matrices: $Q_i$ and $K_i$ with dimensions $L \times d_k$, and $V_i$ with dimensions $L \times d_v$. These matrices are generated by applying linear transformations using weight matrices $W_Q^i$, $W_K^i$, and $W_V^i$, each with dimensions of $d_{emb} \times d_k$ and $d_{emb} \times d_v$, respectively, to the input sequence $X$ for each attention head $i$.

In our specific case, we set $d_k$ and $d_v$ to be equal to $d_{emb} / n_{heads}$, resulting in $d_k = d_v = 384 / 8 = 48$.
To keep the autoregressive property for our model, we mask out the upper right triangle by using the mask matrix $M \in \mathbb{R}^{L\times L}$ shown in Equation \ref{eq:mask_matrix}.
Then, these attention matrices are concatenated with each other along the $d_v$-dimension. 
Afterward, the resulting concatenated matrix is further transformed using another learnable weight matrix $W_O \in \mathbb{R}^{h \cdot d_v \times d_{emb}}$.

\begin{align}
\text{MMHA}(X) &= \text{Concat}(\text{head}_{1}, \text{head}_{2}, \ldots , \text{head}_{h}) \cdot W_O \label{eq:mmha} \\
\text{head}_i &= \text{MaskedAttention}(X \cdot W_Q^i, X \cdot W_K^i, X \cdot W_V^i) \\
\text{{MaskedAttention}}(Q, K, V) &= \text{{softmax}} \left(\frac{{QK^T}}{\sqrt{d_k}} + M \right)\cdot V \\
M &=  \overbrace{\begin{pmatrix}
    0 & -\infty & -\infty & \ldots &-\infty \\
    0 & 0 & -\infty &\ldots &-\infty \\
    \ldots &\ldots & \ldots& \ldots & -\infty \\
    0 & 0 & 0 & \ldots &0 
\end{pmatrix}}^{L} \label{eq:mask_matrix} \\
\text{FFN}(X) &= (\text{SiLu}(X \cdot W_1) \odot (X \cdot W_3)) \cdot W_2 = SwiGLU(X,W_1,W_3) \cdot W_2  \label{eq:ffn}
\end{align}

The Llama 2 architecture employs several changes compared to the standard decoder architecture \cite{vaswani2017attention}.
Firstly, we use rotary positional embeddings (RoPe) \cite{su2022roformer} to encode absolute and relative positional information directly into the attention matrix. Secondly, instead of applying layer normalization \cite{ba2016layernorm} after the self-attention and feed-forward layers, we employ RMSNorm \cite{zhang2019rmsnorm} as a more efficient pre-normalization step. 
A feed-forward layer is described by eq. \ref{eq:ffn}, where $W_1, W_3 \in \mathbb{R}^{d_{emb} \times d_{ffn}}$ and $W_2 \in \mathbb{R}^{d_{ffn} \times d_{emb}}$ are learned weight matrices and $\odot$ represents the elementwise product of two vectors. 
After each feed-forward layer we employ a dropout-layer \cite{dropout} with the probability given in table \ref{table:hyperparams_arch}.

The concat function, which we use below, is defined as follows:
\[
\text{Concat}(A, B) = \left[ \begin{array}{c}
A \\
B \\
\end{array} \right]
\]
Here, $A$ and $B$ are matrices of shape $a \times e$ and $b \times e$ respectively. The result of the concat function is a matrix of shape $(a+b) \times e$, where the rows of $A$ are stacked on top of the rows of $B$.

Furthermore, we made significant alterations to the context ingestion process. The input to our model is a sequence $X$ of shape $L \times d_{emb}$, which can be divided into two parts: $X = \text{Concat}(C,S)$. The first part, $C \in \mathbb{R}^{c \times d_{emb}}$, also later referred to as the \enquote{context}, represents the given conditions and can be expressed as $C = Concat((t_{1}, t_{2}, \ldots, t_{n} )^T, t_{ts}) \in \mathbb{R}^{c \times d_{emb}}$.
The embedded vectors $ t_{i} \in \mathbb{R}^{d_{emb}} \ \forall i \in \{1,\ldots,n\} $ represent the $n$ numerical conditions, which are provided to the model, in our case $n=3$.
On the other hand, $t_{ts} \in \mathbb{R}^{k \times d_{emb}}$ is a matrix of $k$ embedded tokens, that is concatenated with the numerical conditions.
The second part, $S \in \mathbb{R}^{s \times d_{emb}}$, just describes the molecule, as a \acrshort{smiles}, itself.
$c$ is just the length of the complete context and $s$ is the length of the given \acrshort{smiles}, both are not fixed in length.
Typically, a token sequence includes multiple tokens, which translates to multiple embeddings in the context, while a numerical condition is represented by one embedding.
In our case, the order was the following: First are the embedded numerical conditions, then the token sequence embeddings, and lastly the \acrshort{smiles} itself. 
During the training process, we learn \acrshort{smiles} embeddings by learning an embedding vector for each token.

In order to facilitate controlled property generation of molecules, we prepend the sequence with conditions, such as numerical values or a token sequence. While the number of conditions theoretically has no limit, 
we limit the contexts to three numerical values and one token sequence for the purposes of this paper. Each numerical value is assigned a type identifier, and a separate linear layer is used to transform them into the embedding dimension. The transformed values are then combined with the learned type encoding specific to each numerical property. In our implementation, we assigned a fixed type number to each property and mapped it to a learnable vector, which serves as the type encoding.

To provide positional information, we applied RoPe to every part of the context and sequence. Although adding positional information to numerical values is not necessary, we chose to include it for the sake of simplicity in implementation, without negatively impacting the model's performance.

Due to the type identifiers, this approach enables the model to differentiate between various conditions in a straightforward, yet effective manner. Consequently, we are free to mix or even omit conditions within the sequence. This property 
plays a crucial role in our training procedure, as in combination with the SCL method it allows the model to adapt dynamically and process all possible combinations of context.

The degree of creativity of the model's output can be controlled by the so-called temperature parameter, which is defined as a positive real number $t \in \mathbb{R}_{+}$, by dividing the output log probabilities by the said value.
A temperature of $t = 1$ does not alter the model's output, whereas a lower temperature sharpens the output distribution, thus making it more deterministic. Conversely, a temperature greater than one leads to a higher level of variability 

\begin{table}
    \begin{minipage}{0.6\linewidth}
		\centering
		\includegraphics[scale=0.35]{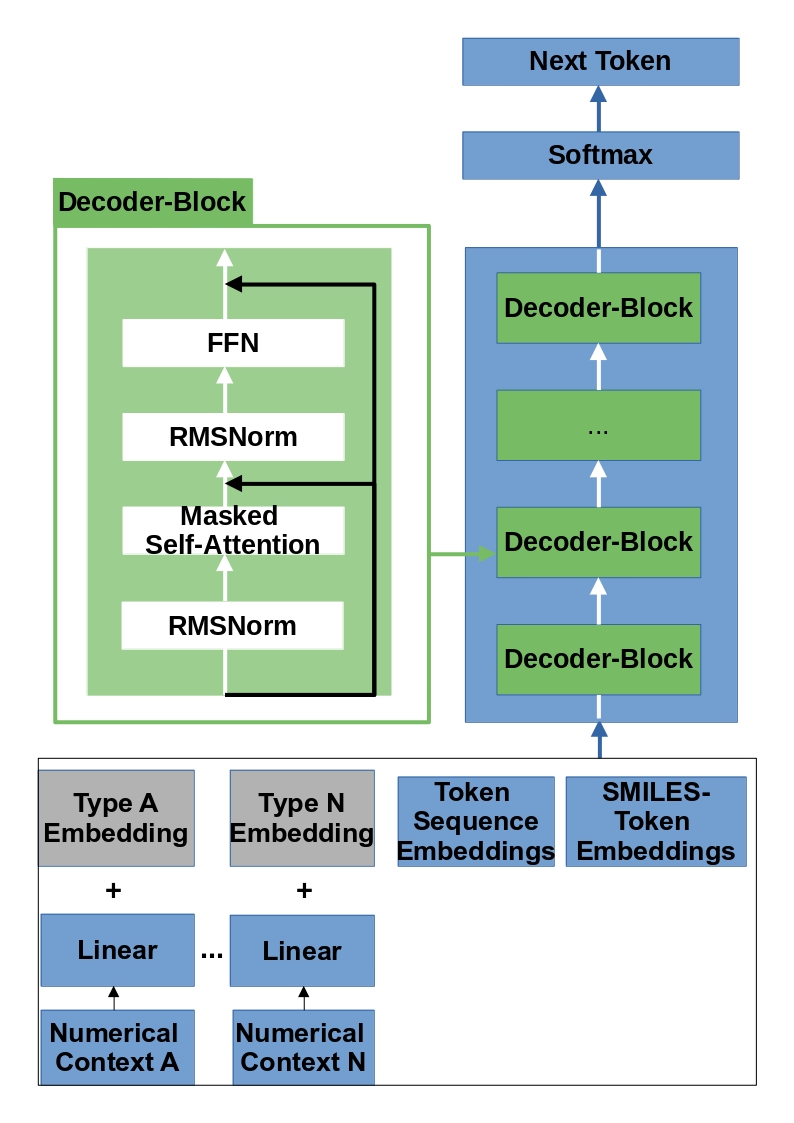}
        \captionof{figure}{The LLamol architecture visualized.} 
		\label{fig:current_architecture} 
 \end{minipage}
	\begin{minipage}{0.4\linewidth}
		\centering
        \caption{Hyperparameters used for the LLamol model.}\label{table:hyperparams_arch}
		\resizebox{\textwidth}{!}{%
		\begin{tabular}{ |p{4cm}||p{1.25cm}|}
 \hline
 \multicolumn{2}{|c|}{Hyperparameters} \\
 \hline
 Parameter / Model & LLamol \\
 \hline
 Number of Attention-Heads $n_{heads}$  & 8     \\
 Number of Decoder-Blocks   & 8     \\
 Dropout Probability   & 10\%     \\
 Activation Function & SwiGLU  \\
 Positional Embeddings & RoPe \\
 Embedding Dimension $d_{emb}$ &   384   \\
 FFN Hidden Dim $d_{ffn}$ &   1024  \\
 Vocabulary Size $d_{voc}$ & 591 \\
 Max \acrshort{smiles} Length & 256 \\
 \hline
\end{tabular} }
    
	\end{minipage}\hfill
\end{table}




\section{Training}

\subsection{Dataset} \label{sec:dataset}

The model was trained on a dataset of molecules, 
which was compiled from several public sources to create a large and diverse population. The resulting dataset, we call \textit{OrganiX13}, includes \acrshort{smiles} strings of mostly organic and/or druglike molecules taken from the 
sources listed in Table \ref{table:dataset_compare}.
\begin{table}[h]
\centering
\caption{Datasets used in the combined dataset.}\label{table:dataset_compare}
\begin{tabular}{||c c||} 
 \hline
 Dataset & Number of SMILES \\  
 \hline\hline
 ZINC 15 \cite{zinc15} & 5M  \\ 
 QM9 \cite{qm9_Ramakrishnan_2014, qm9_Ruddigkeit_2012} & 134k \\
 PC9 \cite{pc9} & 78k \\
 ZINC 250k \cite{zinc250k} & 250k \\
 RedDB \cite{Sorkun_2022} & 31k  \\
 OPV \cite{St_John_2019} & 91k  \\
 PubchemQC 2017/2020 \cite{pubchemqc2020, pubchem2017} & 5.3M  \\
 CEP \cite{cep_harvard} subset \cite{cep_20k}  & 20k \\
 ChEMBL \cite{chembl31} & 2.3M \\
\hline\hline
 Combined (OrganiX13) & 13.1M \\
\hline
\end{tabular}

\end{table}

The original \acrshort{smiles} were left unchanged, however, entries that could not be parsed properly by RDKit were removed. Likewise,
all molecules exceeding a limit of 256 tokens or ionic structures (salts) were rejected.
The final dataset contains approximately 13.1 million \acrshort{smiles}. The potential presence of duplicate and/or synonymous entries was tolerated. 

Subsequently, we used RDKit to provide the numerical values for some quick-to-compute surrogate properties to investigate the training behavior and to enable the direct verification of the generated results.
The properties chosen were the logP, SAScore, and molecular weight, as those properties also have an impact on the achievable energy density or cost of an electro-active material in the chosen aqueous flow battery application. In detail, 

\begin{enumerate}
    \item \textbf{LogP} is defined as the logarithm of the partition coefficient, which denotes the hydrophobic or lipophilic nature of a molecule. 
    A positive logP value suggests that the molecule prefers non-polar solvents, whereas a negative value indicates that the molecule is soluble in water, a desirable property for aqueous flow battery systems, which correlates to energy density.  
    
    \item \textbf{Molecular weight} 
    can be used as a proxy for its size. 
    Again, to attain high energy densities, we would like to have control over the maximum size of the compounds generated. To ensure numerical similarity with the intervals of other properties, the molecular weights were divided by 100. 

   
    \item \textbf{SAScore}: The \acrfull{sascore} \cite{sascore} 
    estimates the  ease or difficulty of creating a compound \cite{ertl2009estimation}. 
    Based on a frequency analysis of chemical moieties in the PubChem database, it assigns a score ranging from zero (easy) to ten (difficult), which is supposed to reflect to some extent the cost of production. 
\end{enumerate}

The resulting dataset encompasses many \acrshort{smiles} strings that cover a broad range of about 12 units in logP, a range in SAScore from around 1 to 6, as well as a similar range in scaled molecular weights. This served as a basis for the subsequent training. 


\subsection{Procedure}

Initially, we convert the \acrshort{smiles} representation into a sequence of tokens using a tokenizer. 
We used the BERT-tokenizer \cite{Schwaller2019_Tokenizer} in DeepChem \cite{Ramsundar-et-al-2019-DeepChem}, which employs a fixed vocabulary size of 591 tokens. 
It splits the \acrshort{smiles} at the character level, except for values enclosed in square brackets, which are treated as a single token.

These tokens are then passed through a separate lookup table, which maps them to a $d_{emb}$-dimensional embedding space. Prior to feeding the token embeddings into the decoder model, a context is added at the beginning.

For each numerical property, we projected the values 
into the embedding space using their respective linear layer and then combined them with the embedded type identifier. 
These identifiers allow the model to distinguish the significance of each numerical value, enabling easy manipulation of their positions or exclusion. Since the properties remain constant throughout the training process, we calculate them in advance.

If we use a token sequence as context, we perform these calculations dynamically in each batch during the training, allowing them to have varying token sequence sizes and content. 
During a training step, a token sequence represents a contiguous subsequence of the current tokenized \acrshort{smiles}. We start by randomly selecting a starting index from zero up to the current \acrshort{smiles} length, followed by determining a random ending index greater than the starting index but smaller than the current \acrshort{smiles} length.

In our case, we limited the context token to a maximum sequence length of 50 to avoid memory issues, which sufficed for our purposes. 
This sequence is then embedded using the same embedding layer as the input sequence and combined with an embedding specific to the token sequence, sharing the shape of the input embedding table. 
Additionally, a learned label embedding is added to these combined token sequence embeddings to indicate their relatedness.

\subsubsection{Stochastic Context Learning (SCL)}

Given an input sequence $X \in \mathbb{R}^{L \times d_{emb}}$ of length $L$, where each element is represented by a $d_{emb}$-dimensional vector, we divide it into two parts: $X = \text{Concat}(C,S)$. 
Our algorithm focuses on modifying the context part $C$. We represent this part as a combination of two parts. The first is $C_{num} = (t_{1}, t_{2}, \ldots, t_{n} )^T \in \mathbb{R}^{n \times d_{emb}}$, where $n$ represents the maximum number of numerical conditions used in the training process (in this case, $n=3$). 
The second is the token sequence $C_{ts} \in \mathbb{R}^{k \times d_{emb}}$, where $k$ is the length of the token sequence, such that $C = Concat(C_{num}, C_{ts})$. The length $k$ is not specific and can change for each input sequence $X$.

To begin, we set a deletion probability $p_{del}$ to $15\%$ during training. For each row in the $C_{num}$ matrix, we check if it should be deleted with a probability of $p_{del}$. 
If it meets the criteria, we remove the row from the $C_{num}$ matrix and consequently from the input sequence $X$, which then would be of shape $(L-i) \times d_{emb}$, where $0 \leq i \leq n $ is the number of deleted numerical conditions.
Similarly, the same probability is used to control if the token sequence should remain in the context for the current sequence.
In this case, the $p_{del}$ probability says if the entire token sequence should be removed, not just one row.
Occasionally, there may be a situation where all conditions in $C_{num}$ and $C_{ts}$ are eliminated. In such instances, the sequence becomes unconditioned.

For batched input sequences $X_{batch}$ with shape $\mathbb{R}^{B \times L_{max} \times d_{emb}}$, the process works similarly. We iterate over each of the $n$ numerical conditions and sample if it should be deleted with a probability of $p_{del}$. If a condition is selected for deletion, we remove the corresponding row from all entries in the batch of size $B$. 
A description for the batched algorithm is given in the Algorithm \ref{alg:scl_alg}.
We assume that every molecule in the batch has all $n$ numerical properties. If a molecule only has a portion of the properties, we would simply pad the missing values. In our case, there was no need for padding, as all molecules had all the numerical properties. 
The batch is created out of $B$ number of sequences $X$, each of those could have a different length $L$ due to the variance in length in the token sequence condition and also the \acrshort{smiles} itself. 
To batch those together, we take the maximum sequence length $L_{max}$ for all sequences that should be packed into the batch and pad the shorter \acrshort{smiles} by appending a pad-token to the length of $L_{max}$.

Thus, throughout the training process, the model has to handle different combinations of the provided conditions, which allows the model to learn unconditionally, single conditions, and also multiple conditions in one go. 
Thanks to the type of embeddings we add to every context element, we can change the number of properties that are provided to the model and still have the model distinguish which properties are provided.

\begin{algorithm}
\begin{algorithmic}[1]
\Require Input sequence $X \in \mathbb{R}^{B \times L \times d_{emb}}$, $n$ maximum number of numerical conditions, B batch size, L sequence size and $d_{emb}$ is the embedding size.
\Function{SCL}{$X$}
    \State $p_{del} \gets 0.15$ \Comment{Deletion probability}  
    \For{$i \gets 1$ to $n$}
        \State $r_1 \gets$ \Call{Random}{$0, 1$} \Comment{Random number between 0 and 1}
        \If{$r_1 \leq p_{del}$}
            \State Remove row $j$ in the L-axis from all samples in $X$ \Comment{Delete numerical condition $j$}
        \EndIf
    \EndFor
    \State $r_2 \gets$ \Call{Random}{$0, 1$} \Comment{Random number between 0 and 1}
    \If{$r_2 \leq p_{del}$}
        \State Remove all entries from the token sequence in the L-axis from all samples in $X$ \Comment{Delete token sequence condition}
    \EndIf
    \State \Return $X$ \Comment{Modified input sequence}
\EndFunction
\end{algorithmic}
\caption{Batched SCL algorithm} \label{alg:scl_alg}
\end{algorithm}

\subsubsection{Loss}
The model is trained to predict the next token by calculating the cross-entropy loss between the actual next token and the predicted probability for that token. 
Note that this loss is only calculated for the \acrshort{smiles} part of the given sequence, the prepended context is not considered in the loss.
Since we only train with the autoregressive loss, the context does not have to be evaluated while training, making our approach very flexible to various conditions.
This loss is then backpropagated through the model using the Adam optimizer \cite{kingma2017adam}.
The cross-entropy loss is defined as follows (Equation \ref{eq:full_formulation_cross_entropy}):
\begin{align}
    \text{CrossEntropyLoss}(y, \hat{y}) = -\frac{1}{N}\sum_{n=1}^{N}\log\left(\frac{\exp(\hat{y}_{n, y_n})}{\sum_{i=1}^{d_{\text{voc}}}\exp(\hat{y}_{n, i})}\right)
     \label{eq:full_formulation_cross_entropy}
\end{align}

In this expression, $N$ is the batch size, where $y \in \{0, 1, 2, \ldots, d_{\text{voc}}\}^{N}$ and $\hat{y} \in \mathbb{R}^{N \times d_{voc}}$ correspond to the target tokens and the predicted log probabilities, respectively.
The mean over the negative logarithms for the normalized predicted probabilities of the next token is calculated. 
Here, $\hat{y}_{n, y_n}$ specifically refers to the predicted log probability assigned to the correct target token $y_n$ for the $n$-th sample in the batch.

The model was trained on a single Nvidia A100 GPU for two days and used about 35 GB VRAM while training. 
A constant learning rate of $\alpha = 10^{-4}$, with $\beta_1 = 0.9$ and $\beta_2 = 0.95$ was used for the Adam optimizer.
The dataset was randomly partitioned into two parts, a training set and a testing set. 
The training dataset consisted of 90\%, while the testing dataset comprised 10\% of the data.
The model was trained using a batch size of 256 with gradient accumulation steps of 4 batches.
Each sequence for the model starts with a \enquote{start of \acrshort{smiles}}-token ([CLS]) and ends with an \enquote{end of \acrshort{smiles}}-token ([SEP]). Shorter \acrshort{smiles} strings were padded with a \enquote{pad}-token ([PAD]) to match the length of the longest \acrshort{smiles} in that batch. The same padding process was applied to the token sequence in the context.

New \acrshort{smiles} 
are then sequentially generated by first starting with a \enquote{[CLS]}-token and then predicting the next tokens iteratively. The generation ends, when the model predicts the \enquote{[SEP]}-token or a specified token limit is reached.

\section {Results and Discussion} 

After the training, we used the model in different scenarios to generate new \acrshort{smiles}, e.g., without any constraints or with one or more constraints (including numerical and/or structural targets), while keeping the temperature parameter constant at $temperature=0.8$. This value ensures a close but not too strict coupling to the underlying probability distributions, which proved helpful in our experiments. 

The metric used to measure the performance of the models for a batch of generated compounds is the mean absolute deviation between requested and obtained numerical values, in addition to the percentage of novelty, uniqueness, and validity of the molecular structures generated. The latter refers to a randomly chosen subset of 2.5 M samples from the training dataset since evaluation against the full dataset proved too much effort. 

In more detail, these metrics are defined as follows:
\begin{enumerate}
    \item \textbf{Novelty}: is defined as the percentage of newly generated molecules not present in the reference 
    dataset. We use this, to ensure that the model is not memorizing the training data, but instead is inventing new compounds. We measure this by comparing the generated \acrshort{smiles} with the \acrshort{smiles} in the dataset. Please note: this is not equivalent to testing the molecular graphs for isomerism, i.e., alternative synonyms are not detected as redundant molecular structures by this procedure, but rather just a string comparison. 
    
    \item \textbf{Uniqueness}: The uniqueness is the ability of the model to generate unique molecules. We measure the percentage of unique molecules generated in a batch of 1k and 10k molecules under specific conditions. 
    Again, identical molecules with synonymous \acrshort{smiles} remain undetected.
    
    \item \textbf{Validity}: The ratio of validity is determined by the number of properly parsed \acrshort{smiles} (by RDKit \cite{rdkit}) versus the total number of generated \acrshort{smiles} in a batch.
    
   
    \item \textbf{Mean average deviation}: Is defined as the following:
    \begin{align}
        MAD &= \frac{1}{n} \sum_{i=1}^{n}{ |x_i - y_i|}
    \end{align}
    
    For each of the $n$ generated \acrshort{smiles} strings, the target value of the respective property is denoted as $y_i$, while $x_i$ represents the 'true', i.e. actually calculated property value.  
    The model should minimize this quantity without being explicitly trained on it, which would indicate that the model incorporates the provided context into the generative process.
    This metric is also used to enable comparisons to other models, e.g. \cite{molgpt}.
    
\end{enumerate}

We specifically compare our model to MolGPT \cite{molgpt} as it is the most similar in terms of architecture and choice of conditions.

\subsection{Unconditional Generation}

\begin{figure}[h]
\centering
\begin{subfigure}[b]{0.33\textwidth}
    \centering
    \includegraphics[width=\linewidth]{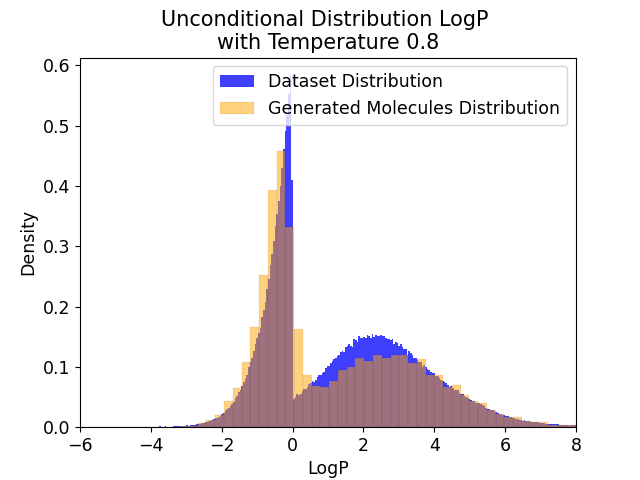}
    \caption{LogP}
    \label{fig:compar_unc_data_logP}
\end{subfigure}
\hfill
\begin{subfigure}[b]{0.33\textwidth}
  \includegraphics[width=\linewidth]{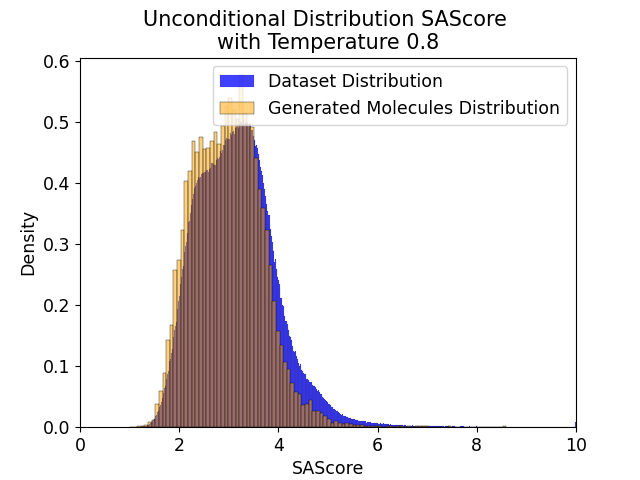}
  \caption{SAScore}
  \label{fig:compar_unc_data_sascore}
\end{subfigure}
\hfill
\begin{subfigure} [b]{0.33\textwidth}%
  \includegraphics[width=\linewidth]{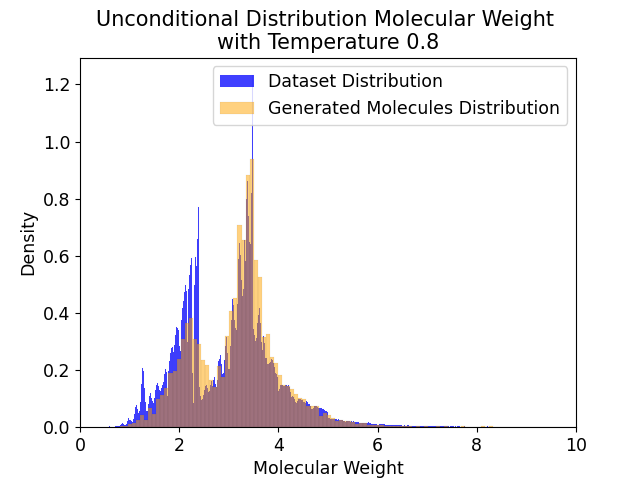}
  \caption{Molecular weight}
  \label{fig:compar_unc_data_molweight}
\end{subfigure}
\caption{Distribution of properties as obtained from a 2.5M sample of training molecules in comparison with the distributions from 20k unconditionally generated molecules.}
\end{figure}

Without applying any conditions, we generated 20k \acrshort{smiles} and calculated the corresponding properties logP, SAScore, and molecular weight using RDKit. The resulting frequencies of distribution are very similar to the distributions obtained from a representative sample of training molecules,  
see Figure \ref{fig:compar_unc_data_logP}, 
\ref{fig:compar_unc_data_sascore}, and 
\ref{fig:compar_unc_data_molweight}. 
This indicates that the model has indeed learned the inherent 
distribution of the training dataset, without specifically training the model unconditionally. 

The generated \acrshort{smiles} also achieve very comparable performance in terms of uniqueness, and validity to MolGPT as shown in Table \ref{table:metrics_compare} (row 1 and 8). Surprisingly, our degree of novelty is significantly higher than in the MolGPT case. We suspect that this is mostly due to our larger dataset, which makes it much more unlikely for the model to reproduce a specific entry. 


\hfill

\subsection{Single Condition}

In this experiment, we also assessed the model's ability to handle single-condition generation over wide ranges of target values. 

For each target value of the intervals listed in Table \ref{table:metrics_compare}, rows 2 -- 7, the procedure involved generating a sample of 10,000 molecules. 
As before, we determined the true property values of the generated molecules using RDKit and compared those to the targeted values.
For each property, we ran two scenarios: 
The first one covered a broad interval of values that encompassed both in-distribution, as well as out-of-distribution values. 
The second run focused solely on the performance of a select few in-distribution target values.
The interval notation [$a$, $b$; $c$] signifies that the values were uniformly sampled from a discrete distribution of values, including $a$ and $b$, with a step size of $c$. More precisely, [-4, 7; 0.5] denotes the set of values $\{-4, -3.5, -3, ..., 6.5, 7\}$. The notation $\{ a_1, a_2, ... \}$ indicates that the values were uniformly sampled from that specific set of values.

\begin{figure}[h]
\begin{subfigure}[b]{0.33\textwidth}
  \includegraphics[width=\linewidth]{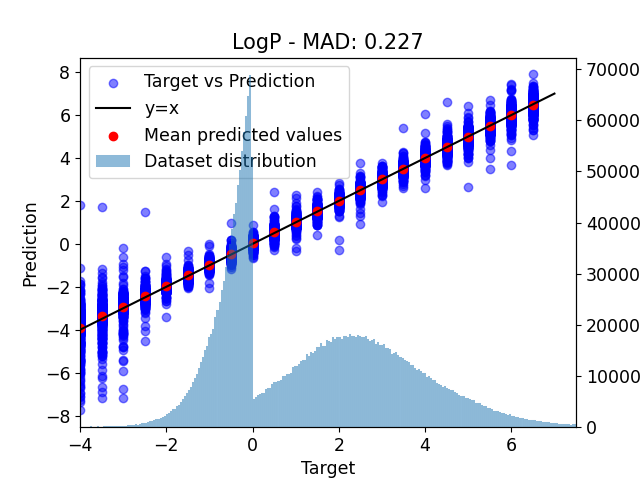}
  \caption{logP}\label{fig:sing_cond_logP}
\end{subfigure}
\hfill
\begin{subfigure}[b]{0.33\textwidth}
  \includegraphics[width=\linewidth]{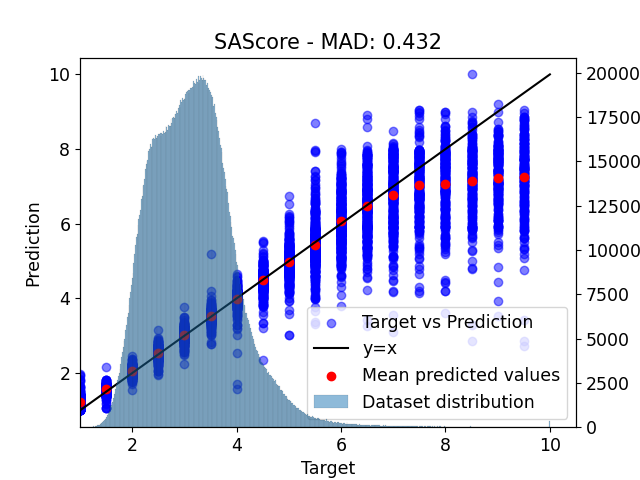}
  \caption{SAScore }
  \label{fig:sing_cond_sascore}
\end{subfigure}
\hfill
\begin{subfigure}[b]{0.33\textwidth}%
  \includegraphics[width=\linewidth]{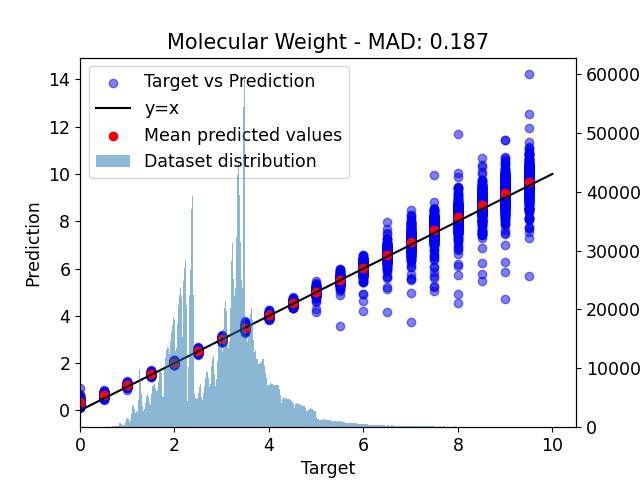}
  \caption{molecular weight }
  \label{fig:sing_cond_molweight}
\end{subfigure}

\caption {Requested (x-axis) versus actual values (y-axis) for the diverse target properties: a) logP, b) SAScore, c) molecular weight. For each target value, a batch of 10k \acrshort{smiles} was generated; MAD is averaged over the entire range. }
\end{figure}

Despite the low probability of the model being trained solely on one property, it performs 
well in this task, as demonstrated in Figures \ref{fig:sing_cond_logP}, \ref{fig:sing_cond_sascore}, and \ref{fig:sing_cond_molweight}.
The model achieves 
low MAD values across the entire span of the respective target properties (rows 2, 4, and 6), although the MAD values obtained for the in-distribution series are generally and expectedly significantly lower (rows 3, 5, and 7).   
In fact, predicting logP values to an accuracy of 0.5 logP units (root mean square deviation) is commonly considered a satisfactory result \cite{icsik2020assessing}.


Still, the out-of-sample performance is acceptable
. Although the scatter increases, the general trend is well retained, with the only exception of the very high (>7) SAScores. In this case (row 4 in Table \ref{table:metrics_compare}), we observe also a concomitant drop in the percentage of validity of the generated molecules (81 vs 99.7 \%). Upon manual inspection, we find that not only compounds with highly bridged ring systems and/or accumulations of stereogenic centers were generated that are supposedly very demanding to synthesize but also rare and unstable atomic environments such as neighboring diradicals and/or carbenes that presumably prevent the proper parsing of structures. 

In comparison to MolGPT our model archives a slightly lower MAD in the single condition case, without being specifically trained on that task, while simultaneously scoring the same uniqueness and validity.

\begin{table}[h]
\centering
\caption{Table for comparing metrics for the three metrics at a temperature of 0.8 for 10k generated molecules. All metrics are evaluated with these 10k molecules, except uniqueness at 1k.}
\label{table:metrics_compare}
\begin{tabular}{||l p{0.1\textwidth} p{0.1\textwidth} p{0.1\textwidth} p{0.1\textwidth} p{0.1\textwidth} p{0.075\textwidth}  p{0.075\textwidth}||} 
\hline
Model &Condition type & Interval & Novelty [\%] $\uparrow$ & Uniqueness @ 1k [\%] $\uparrow$& Uniqueness [\%]$\uparrow$ & Validity [\%] $\uparrow$  & MAD $\downarrow$\\

\hline\hline
Llamol & Unconditional & & \textbf{97.58}  & \textbf{100.0}  & \textbf{100.0} & \textbf{99.49} & \\
& LogP & [-4, 7; 0.5] & 97.74  & 100.0  & 99.72 & 99.57 & 0.227  \\
& \textbf{LogP} & \{2, 4, 6\} & 97.45  & 100.0  & \textbf{99.82} & \textbf{99.61}  &  \textbf{0.194}  \\
& SAScore & [1, 10; 0.5] & 97.09  & 99.1  & 97.07 & 81.23  &  0.432 \\
& \textbf{SAScore} & \{2, 3, 4\} & 97.41  & 100.0  & \textbf{99.94} & \textbf{99.7}  &  \textbf{0.099} \\
&  Molecular weight& [1, 10] & 97.84  & 99.8  & 99.51 & 97.33 & 0.187 \\
&  Molecular weight& \{2, 3, 4\} & 97.37  & 100.0  & 99.97 & 99.52 & 0.041 \\
\hline 
MolGPT & Unconditional &  & 79.7  & 100.0 & 100.0 & 99.4 &   \\
& LogP & \{2, 4, 6\} & 100.0  &   & 99.8 & 97.1 & 0.23  \\
& SAScore & \{2, 3, 4\} & 100.0  &   & 99.5 & 97.7 & 0.13  \\
\hline
\end{tabular}

\end{table}



\subsection{Multiple Conditions}

\begin{figure}[h]
\captionsetup{margin=5pt}
\minipage{0.5\textwidth}
  \includegraphics[width=\linewidth]{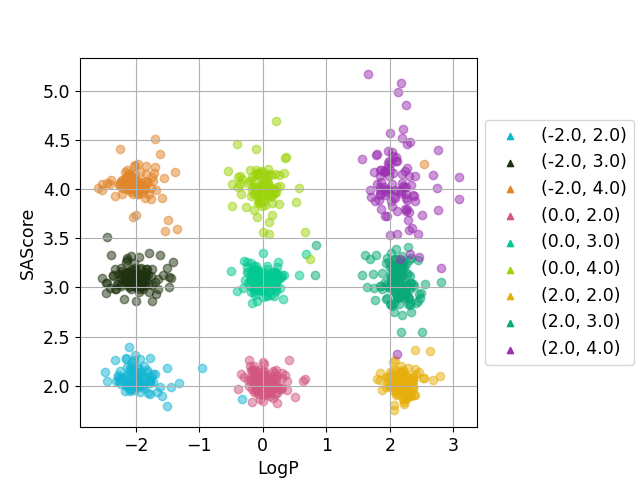}
  \captionsetup{skip=5pt} 
  \caption{logP + SAScore}\label{fig:mult2_cond_logP}
\endminipage\hfill
\minipage{0.5\textwidth}
  \includegraphics[width=\linewidth]{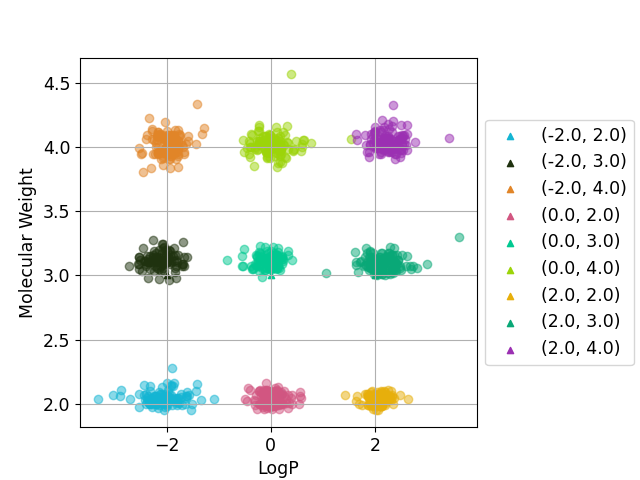}
  \captionsetup{skip=5pt} 
  \caption{logP + molecular weight}\label{fig:mult2_cond_sascore}
\endminipage\hfill
\minipage{0.5\textwidth}%
  \includegraphics[width=\linewidth]{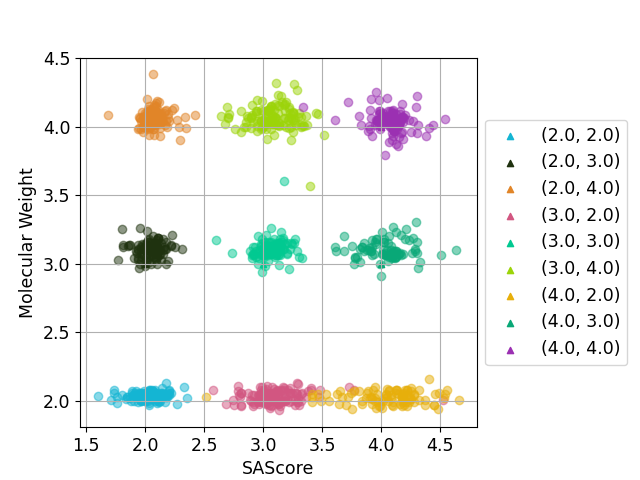}
  \captionsetup{skip=5pt} 
  \caption{SAScore + molecular weight}\label{fig:mult2_cond_molweight}
\endminipage
\minipage{0.5\textwidth}%
  \includegraphics[width=\linewidth]{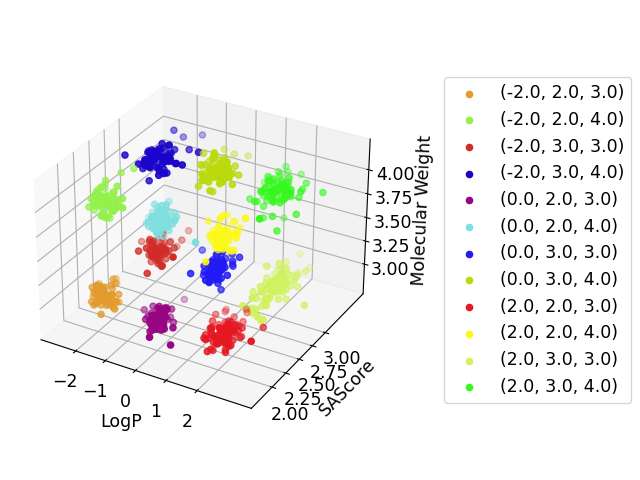}
  \captionsetup{skip=5pt} 
  \caption{logP + SAScore + molecular weight}\label{fig:mult3_cond_lsm}
\endminipage
\end{figure}

For each pairwise combination of target properties, we generated 1k \acrshort{smiles}, see
Figures \ref{fig:mult2_cond_logP}, \ref{fig:mult2_cond_sascore} and \ref{fig:mult2_cond_molweight}.
The graph labels are in the same order as given in the captions of the respective figures.

In general, 
the generated molecules' actual properties center closely around the desired values. 
Although all chosen values were well within the highly populated areas of the underlying distributions, some combinations turned out to be hard to satisfy, 
resulting in a more pronounced scatter. 

Figure \ref{fig:mult2_cond_logP} shows the distribution of calculated logP values and SAScores. 
This pair works well for lower logP values, but for higher ones the variance in the SAScore axis rises significantly. 
This seems to indicate that in this case, the logP values have a slight priority in the generative process compared to SAScore.
There are some outliers, but most of the generated molecules fulfill both conditions.
We suspect that this could be an effect of the shortage of training data in that region, thus leading to more inaccurate results. 

Next we compare the combination of logP and the molecular weight values, as shown in Figure \ref{fig:mult2_cond_sascore}. 
Apparently, the molecular weight takes priority in the generation, as it displays a much smaller variance compared to the logP. 
However, the logP is still met accurately despite being under very strict size constraints. 
This comes as no surprise, due to the ease with which the molecular weight can be determined by counting the contributions of each atom, as opposed to the more extensive considerations demanded by logP values.

Lastly, Figure \ref{fig:mult2_cond_molweight} displays the combination of SAScore and molecular weight. Similar to the logP and molecular weight comparison, molecular weight still dominates the generative process. In comparison, the model can not uphold the SAScore in all cases. 
This is especially evident in the case, where the molecular weight is set to a low value of 1.5 which results in a high SAScore variance. 
Apparently, the model struggles to incorporate a sufficient number of challenging motifs into a small molecule, due to the limited size and range of available elements
In contrast, when the weight is set to a higher value of 3.5, i.e. a larger molecule, we obtain a much lower variance.

Finally, in Figure \ref{fig:mult3_cond_lsm} we visualize the generated molecules that take into account all three properties. 
As is evident by the disjoint point clouds in the graph, the model learned to consider 
all three conditions and generate matching molecules. 
The labels in the graph should be read in the order of logP, SAScore and then molecular weight.

In Table \ref{table:model_comp_multicond} we compare the performance of our model to MolGPT in multi-conditional generation where applicable.
Each row in the table represents an experiment, with the
columns representing the properties used as conditions. If a condition was not utilized, the cell was left empty.
Our model has on-par performance compared to MolGPT in the case of logP + SAScore, while simultaneously being able to handle other condition combinations effectively.


\begin{table}[h]
\centering
\caption{Table for comparing multiple property conditions for 10k generated molecules to other models.}
\label{table:model_comp_multicond}
\begin{tabular}{||p{0.10\textwidth} p{0.10\textwidth} p{0.10\textwidth} p{0.10\textwidth} p{0.12\textwidth} p{0.12\textwidth} p{0.12\textwidth}||} 
\hline
Model & Novelty [\%] $\uparrow$ & Uniqueness [\%]$\uparrow$ & Validity [\%] $\uparrow$  &  LogP \{2, 4, 6\} MAD $\downarrow$ & SAScore \{2, 3, 4\}  MAD $\downarrow$ & Molecular Weight \{2, 3, 4\} MAD $\downarrow$\\  
\hline\hline
Llamol & 98.1 & 99.5 & 99.1 & \textbf{0.21} & 0.15 & \\
& 97.6 & 89.3 & 99.1 & 0.21 &  & 0.04\\
& 97.4 & 99.4 & 99.3 & & 0.11 & 0.04\\
& 98.2 & 95.4 & 98.6 & 0.26 & 0.19 & 0.04\\
\hline 
MolGPT & 100.0 & 99.2 & 97.2 & 0.25 & 0.14 &  \\
\hline
\end{tabular}

\end{table}

\subsection{Token Sequence Incorporation}

A very common question in material design is to create analogs from a given starting molecule and add/modify structural features to customize the physical properties. For this reason, the model also accepts a \acrshort{smiles} string representing the desired molecular moiety that should be integrated as a building block in the newly generated structure as an additional condition.
To measure the performance of our model for this task, we used the following criteria: 
\begin{itemize}
    \item \textbf{Substructure Matches (SM)}: The substructure match measures the percentage of generated molecules that explicitly include the target moiety. 
    As a first step, we convert the target structure into a SMARTS \cite{smarts} pattern, which is essentially a regular expression to match specific atoms or substructures within a molecular structure. 
    To make the criterion a little less strict, all information about bond orders is removed, only the connectivity itself is retained. Therefore, the pattern tolerates modifications in the details of the electronic structure (e.g. localized double bonds versus aromatic bonds), while still maintaining the overall topology.  
    With this property, we measure how often the target structure is retained during the generative process.
\end{itemize}

For this experiment, at a constant temperature of 0.8, batches of 1000 molecules were generated for various context token sequences and evaluated using the mentioned metric. Table \ref{table:metrics_compare_molfragments} lists different organic target structures (as the context token sequence in \acrshort{smiles} form) and the results obtained a) without applying any other conditions (columns: uniqueness at 1k / SM), and b) with another additional numerical condition (columns: LogP/SAScore/molecular weight at different target values each). 
 
Overall, the model seems to perform very well, as we can recover the target structures at least once in most of the newly generated \acrshort{smiles}, except for one particular example, Thiophene, see below.  
However, especially when given a larger target structure like for example Morphine, we observe that the generated structures become very repetitive. 



By chance, we found that the success rate in generating structures containing the building block thiophene is very dependent on the explicit formulation of the \acrshort{smiles} string (rows 2 + 3 in Table \ref{table:metrics_compare_molfragments}).
Thiophene is a common hetero-aromatic substructure in the training dataset. 
It can be expressed in lowercase letters, which represent aromatic ring systems, or in uppercase letters, indicating localized double bonds. 
The training data mostly use the aromatic notation (in various synonyms, depending on where the listing of ring atoms starts). 
Overall, we found about 144k \acrshort{smiles} strings that contain at least one thiophene substructure. 
Table \ref{table:thiophene_analysis} shows the synonyms, their frequencies in the training data, and the retrieval rates (for a batch size of 100 \acrshort{smiles}).

The recovery rate for this target moiety varies significantly, ranging from 10\% to 70\% for various aromatic synonyms. However, it exhibits only a loose correlation with the observed relative frequency of each synonym in the training data. In contrast, a \acrshort{smiles} in kekulized notation—i.e., with localized double bonds, a rarity in the training data—achieves a retrieval rate of 90\% for the specified token sequence. Notably, this sequence is expressed in aromatic lowercase notation in the generated \acrshort{smiles}.
We suspect, that the discrepancies in performance trace back to the internal perception of the target context token sequence due to the rather complicated rules defining aromaticity and the relative position of the tell-tale sulfur atom as a main feature of thiophene. In contrast, the explicit specification of double bonds seems to allow a more reliable (re)construction of the overall structure, despite its almost complete lack in the training set.

\begin{table}[h]
\centering
\begin{tabular}{p{2cm}p{2cm}p{2.5cm}p{2cm}}
\toprule
\textbf{SMILES} & \textbf{Dataset\newline Occurrences} & \textbf{Substructure Match (\%)} & \textbf{Validity (\%)} \\
\midrule
s1cccc1 & 0 & 11 & 91 \\
c1sccc1 & 17k & 66 & 98 \\
c1cscc1 & 7k & 56 & 98 \\
c1ccsc1 & 47k & 68 & 99 \\
c1cccs1 & 73k & 42& 99 \\
C1=CSC=C1 & 288 & 90 & 99 \\
\bottomrule
\end{tabular}
\caption{Thiophene (as a context token sequence) analysis for 100 generated molecules} \label{table:thiophene_analysis}

\end{table}

\begin{table}[h]
\centering
\caption{Table for comparing metrics on 1000 generated molecules for each context token sequence.}\label{table:metrics_compare_molfragments}
\begin{tabular}{||l|p{0.3\textwidth}|p{0.125\textwidth}|p{0.15\textwidth}|p{0.15\textwidth}|p{0.15\textwidth}||}
 \hline
 & Token sequence \acrshort{smiles} & Unconditional {\color{cyan} Uniqueness at 1k [\%]} / {\color{orange} SM [\%]} & LogP \{-2, 0, 2\} MAD / {\color{cyan} Uniqueness at 1k [\%]} / {\color{orange} SM [\%]} & SAScore \{2, 3, 4\} MAD / {\color{cyan} Uniqueness at 1k [\%]} / {\color{orange} SM [\%]} & Molecular Weight \{2, 3, 4\} MAD / {\color{cyan} Uniqueness at 1k [\%]} / {\color{orange} SM [\%]}  \\  
 \hline\hline
 1 & c1ccccc1 (Benzene) & \hspace{0px} {\color{cyan}100.0} / {\color{orange}89.6} & 0.43 / {\color{cyan}99.6} / {\color{orange}69.3} & 0.16 / {\color{cyan}99.9} / {\color{orange}85.0}  & 0.1 / {\color{cyan}98.8} / {\color{orange}88.8} \\
 2 & s1cccc1 (Thiophene) & \hspace{0px} {\color{cyan} 96.5}  / {\color{orange}10.7} & 0.54 / {\color{cyan}98.8} / {\color{orange}17.6} & 0.2 / {\color{cyan}93.8} / {\color{orange}32.7}  &  0.15 / {\color{cyan}96.9} / {\color{orange}30.5}  \\
 3 & C1=CSC=C1 (Thiophene) & \hspace{0px} {\color{cyan} 90.1} / {\color{orange}88.98} & 0.48 / {\color{cyan}90.9}  / {\color{orange}87.7}  &  0.60 / {\color{cyan}79.3}  / {\color{orange}93.2}  & 0.08 / {\color{cyan}88.2} / {\color{orange}72.2}\\
 4 & CC1=CSC=C1 (3-Methylthiophene) & \hspace{0px} {\color{cyan} 86.5} / {\color{orange}86.8} &  0.57 / {\color{cyan}81.4} / {\color{orange}92.7} & 0.44 / {\color{cyan}48.6} / {\color{orange}95.2}  & 0.11 / {\color{cyan}78.5} / {\color{orange}75.6}\\
 5 & CCO (Ethanol) & \hspace{0px} {\color{cyan}100.0} / {\color{orange}60.8} & 0.16 / {\color{cyan}99.9} / {\color{orange}65.7} &  0.1 / {\color{cyan}100.0} / {\color{orange}65.3}  &  0.07 / {\color{cyan}99.5} / {\color{orange}63.2} \\
 6 & CC=O (Acetaldehyde) & \hspace{0px} {\color{cyan}91.2} / {\color{orange}87.8} & 0.24 / {\color{cyan}94.0} / {\color{orange}93.7} & 0.2 / {\color{cyan}90.6} / {\color{orange}89.2}  &  0.07 / {\color{cyan}97.6} / {\color{orange}89.3} \\
 7 & CC(=O)OC1=CC=CC=C1C(=O)O (Aspirin) & \hspace{0px} {\color{cyan}51.7} / {\color{orange}97.8} & 0.5 / {\color{cyan}60.7} / {\color{orange}97.1} &  0.17 / {\color{cyan}73.3} / {\color{orange}96.6} &  0.09 / {\color{cyan}28.6} / {\color{orange}97.7}    \\
 8 & CC(=O)NC1=CC=C(C=C1)O  (Paracetamol) & \hspace{0px} {\color{cyan}90.9} / {\color{orange}81.6} &  0.64 / {\color{cyan}75.0} / {\color{orange}75.3}  & 0.19 / {\color{cyan}95.9} /{\color{orange} 78.6}  &  0.08 / {\color{cyan}62.2} / {\color{orange}72.1}\\
 9 & CN1C=NC2=C1C(=O)N (C(=O)N2C)C (Caffeine) & \hspace{0px} {\color{cyan}11.8} / {\color{orange}99.8} &  0.69 / {\color{cyan}36.5} / {\color{orange}96.9}  &  0.5 / {\color{cyan}25.9} / {\color{orange}96.6}  &  0.14 / {\color{cyan}43.9} / {\color{orange}96.3}\\
10 & CN1CCC23C4C1CC5=C2C (=C(C=C5)O)OC3C(C=C4)O (Morphine)& \hspace{0px} {\color{cyan}14.7}  / {\color{orange}99.8}  &  1.05 / {\color{cyan}25.6} / {\color{orange}99.7}  &  2.29 / {\color{cyan}9.5} / {\color{orange}99.8}  &  0.1 / {\color{cyan}27.0} / {\color{orange}95.4} \\ 
11 & OC(=O)C(C)c1ccc(cc1)CC(C)C (Ibuprofen) & \hspace{0px} {\color{cyan}14.2} / {\color{orange}95.4} & 1.42 / {\color{cyan}88.2} / {\color{orange}67.4} & 0.33 / {\color{cyan}53.2} / {\color{orange}84.8}  & 0.1 / {\color{cyan}53.8} / {\color{orange}72.8}
 \\
\hline
\end{tabular}

\end{table}

\subsection{Token Sequence with a Single Numerical Condition}

The really useful application for customizing given structures is the simultaneous application of one or more additional criteria. 

Thus, we study combinations of token sequence conditions together with single numerical conditions, see 
Table \ref{table:metrics_compare_molfragments}. Each combination was tested on 1000 generated molecules, with the numerical values uniformly sampled from the range specified in the table header for each property.

In most cases, we observed a decrease in the number of substructure matches for the molecules tested as compared to the previous run without numerical conditions. 
This is likely due to the model having to handle two possibly competing conditions simultaneously. The MAD values for logP and SAScore were also notably higher compared to generating without a token sequence but remain within acceptable limits. 
It is worth noting that when the two conditions conflicted, such as with Ibuprofen and negative logP values, this led to the presence of some significant outliers. Conversely, when the conditions aligned well, the errors were consistent with the previous results. More details on the graph for the Ibuprofen and logP relationship can be found in Appendix \ref{subsec:appendix_a_ibu}.
We also observed that the SAScore in the case of Morphine is significantly higher than in the other examples. This is mostly due to Morphine having a SAScore of about 5.2, and we requested lower values. 
In this case, the model prioritizes the token sequence in comparison to the SAScore, which leads to the higher MAD.

Apparently, the token sequence condition takes precedence in most cases over the criteria logP and SAScore, as evidenced by the elevated MAD scores. 
Yet again, the molecular weight seems to be prioritized over the token sequence, as evidenced by the very low MAD scores, particularly for larger molecules such as Morphine.

\begin{table}[h]
\centering
\caption{Table for comparing multiple property conditions for 1000 generated molecules using 4 example token sequences.}
\label{table:metrics_multicond2_compare_molfragments}
\begin{tabular}{||p{0.35\textwidth} p{0.075\textwidth} p{0.10\textwidth} p{0.10\textwidth} p{0.10\textwidth} p{0.10\textwidth}||} 
\hline
Token Sequence SMILES &  SM[\%] & Uniqueness at 1k [\%] & LogP \{-2, 0, 2\} MAD & SAScore \{2, 3, 4\}  MAD & Molecular Weight \{2, 3, 4\} MAD \\  
\hline\hline
C1=CSC=C1 (Thiophene) & 80.1 & 79.9 & 0.61 & 0.27 & \\
& 73.4 & 87.8 & 0.47 &  & 0.10\\
& 77.3 & 84.2 &  & 0.25 & 0.09\\
& 68.6 & 80.9 & 0.56 & 0.28 & 0.10\\
\hline
CC=O (Acetaldehyde) &  91.9 & 91.9 & 0.21 & 0.19 &  \\
& 95.3 & 98.7 & 0.20 &  & 0.06 \\
& 87.8 & 97.6 &  & 0.14 & 0.06\\
& 94.6 & 97.7 & 0.24 & 0.15 & 0.06\\
\hline
CC(=O)NC1=CC=C(C=C1)O (Paracetamol) & 82.7 & 81.3 & 0.41 & 0.21 & \\
& 61.7 & 62.1 & 0.62 & & 0.10 \\
& 77.8 & 75.2 & & 0.42 & 0.09\\
& 68.9 & 75.4 & 0.49 & 0.42 & 0.09\\
\hline 
CN1C=NC2=C1C(=O)N(C(=O)N2C)C (Caffeine) & 94.1 & 53.6 & 0.56 & 0.35 &  \\
& 75.5 & 47.6 &  0.55 &  & 0.10 \\
& 88.0 & 37.0 &  & 0.42 & 0.10 \\
& 77.7 & 49.6 & 0.53 & 0.42 & 0.08 \\
\hline
\end{tabular}

\end{table}

\subsection{Token Sequence with Multiple Numerical Conditions}

We also conducted experiments where multiple token sequences were tested under two conditions simultaneously. The results of these experiments can be found in Table \ref{table:metrics_multicond2_compare_molfragments}. Each row in the table represents a specific experiment, with the columns representing the properties used as conditions. If a condition was not utilized, the cell was left empty.

The model consistently performs well under various conditions, as shown by the low MAD values. However, when conditions are overly restrictive in combination with the token sequence, it can lead to higher MAD values. This is because the model prioritizes certain properties over others.

For instance, consider Paracetamol, where both logP and molecular weight conditions are applied. Due to the constraining effect of molecular weight on the molecule's size, decreasing the logP value significantly becomes challenging. In this case, the model prioritizes the molecular weight condition. We suspect this is because molecular weight is easier to validate and has more pronounced limitations compared to logP.

Nevertheless, the model effectively satisfies all three constraints in most cases, as evidenced by a high percentage of substructure matches and low MAD values for the properties in Table \ref{table:metrics_multicond2_compare_molfragments}. Notably, when generating molecules with three properties, some MAD values are even lower than those observed in two-property generation. This could be attributed to the model being trained on a larger number of three-property batches, resulting in improved performance.

In general, all four conditions are respected during the generative process and make significant contributions to the resulting molecules.

\section{Conclusion}


Our aim was to provide a tool for exploring the relevant chemical spaces for a given application, in our case the subspace of organic, potentially electro-active compounds. We therefore adapted existing work and approaches to our needs and came up with a new training variant that allows for a solitary model very flexible in use, which was also trained on a data set of substantial size. 

In detail, we
\begin{enumerate}
    \item developed a GPT-style Transformer based on the LLama 2 architecture, showcasing strong performance in both single and multi-conditioned generation, comparable to or slightly surpassing existing models, despite not being task-specific.
    \item compiled and utilized a training dataset comprising 13 million organic molecules sourced from various origins, enhancing the model's ability to generate a variety of molecular structures.
    \item implemented a new training method we call Stochastic Context Learning (SCL), enabling our model to handle various combinations of conditions efficiently for multi-conditional generation using a single model.
\end{enumerate}

We were able to show that the training process was successful and the achieved accuracy very satisfactory. The model generalizes quite well, as target values requested outside the well-sampled areas still tend to fall in the desired ranges.  

The whole setup is very generic and easily adaptable to other applications.
The latter motivates the number and choice of properties used as conditions for narrowing down the search space. In fact, for the model to be more useful in the search for energy-storage materials, in future we intend to provide a more meaningful, yet expensive property, such as the enthalpy of reaction.


Looking ahead, this research opens up exciting possibilities for further advancements in generative models and their applications in chemistry and related fields. Our modified architecture, combined with the SCL approach, holds great potential for generating novel and diverse organic molecules with precise control over desired properties.

In theory, a single model can learn a wide range of conditions and combinations by utilizing this approach during training.
Therefore, we chose the SAScore (reflecting a materials' production cost), molecular size and logP (contributing to the energy density), as well as a desirable molecular core structure as optional target conditions. As an added benefit, a single model also comes at a reduced training cost.
This method enables a more flexible and scalable training process, as it does not require every property to be available for all samples.

\textbf{Outlook}

For future work, we intend to focus more on curating the dataset, as to not have these very concentrated distributions for all properties. We hope that by reducing redundant molecules, the model would generalize better, while also reducing training time in the process. 
Generally, we assume that the model could perform even better with more training data, as it seems to be underfitted even with our large dataset.

Furthermore, we also intend to expand the number of properties that are given to the model, as there are more useful conditions for practical applications, such as the HOMO-LUMO gap.

\section*{Acknowledgments}
This work was supported in part by the BMBF-project 05M2AAA {MaGriDo} (Mathematics for Machine Learning Methods for Graph-Based Data with Integrated Domain Knowledge) and in part by the project SONAR funded by the European Union’s
Horizon 2020 research and innovation program under Grant Agreement
no. 875489. This paper reflects only the authors' view. The funding
agencies are not responsible for any use made of the information it
contains.

\section*{Code and Data Availability}

The code and dataset are available here: https://github.com/Fraunhofer-SCAI/llamol

\section*{Appendix}

\subsection{A} \label{subsec:appendix_a_ibu}

In this chapter, we visualize the errors of generated molecules using the molecular fragment condition with a single numerical condition. 

\begin{figure}[h]
    \centering
      \includegraphics[scale=0.55]{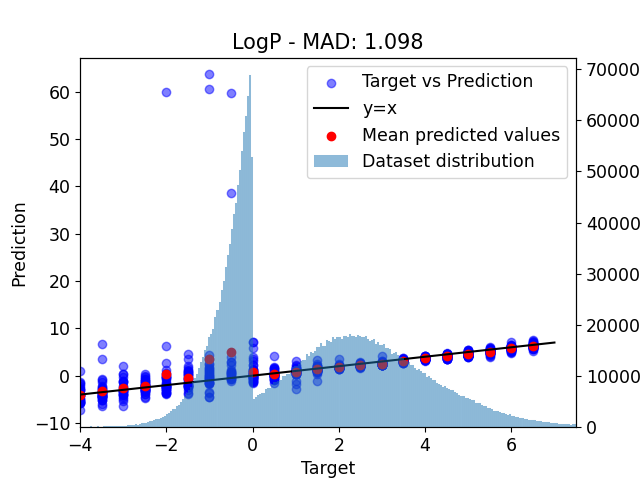}
      \small{\caption{Ibuprofen logP Graph - Generated vs Target} \label{fig:ibuprofen_logp}}
\end{figure}

In the case of Ibuprofen with a naturally very positive logP of about 3.0, it is very difficult for the model to significantly reduce the logP to the desired negative values, while also keeping the fragment intact.
This leads to an overall higher MAD, due to a small sample of large outliers that increased the mean by a significant margin.
This can be seen in the Figure \ref{fig:ibuprofen_logp}.

\subsection{B} \label{subsec:appendix_b_special_cases}

We also conducted some experiments on special combinations of different conditions, as these also show the limitations of the model, either due to the incompatibility of these conditions or the lack of training data in those regions.

\begin{figure} 
    \centering
      \includegraphics[width=1.0\textwidth,height=1.0\textwidth]{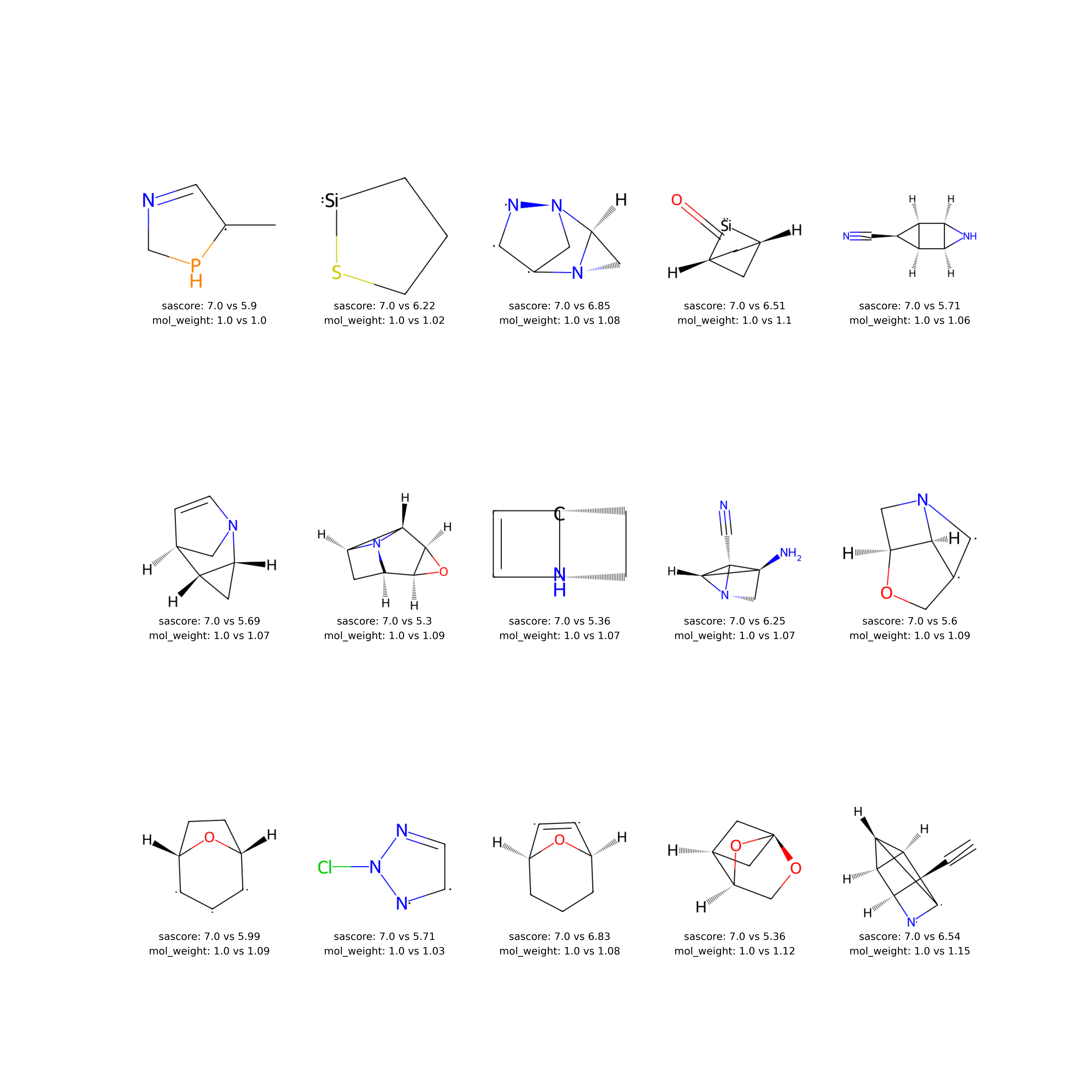}
      \small{\caption{Special Case: Generated molecular with low molecular weight and high SAScore as conditioning.} \label{fig:sc_low_weight_high_sas}}
\end{figure}

We tested the combination of a low molecular weight (100) and a high SAScore (7), which can be seen in the Figure \ref{fig:sc_low_weight_high_sas}.
The generated molecules have both characteristics by being hard to produce due to the high number of connected, bridged, annealed or spiro-rings and ring strains associated with the high degree of interconnected rings and/or open-shell centers (radicals and/or carbenes), while keeping the molecular weight small. 
In this scenario, it also uses more uncommon elements to fit into both conditions. 

\subsection{C} \label{subsec:appendix_c_samples}

\begin{figure} 
    \centering
      \includegraphics[width=1.0\textwidth,height=1.0\textwidth]{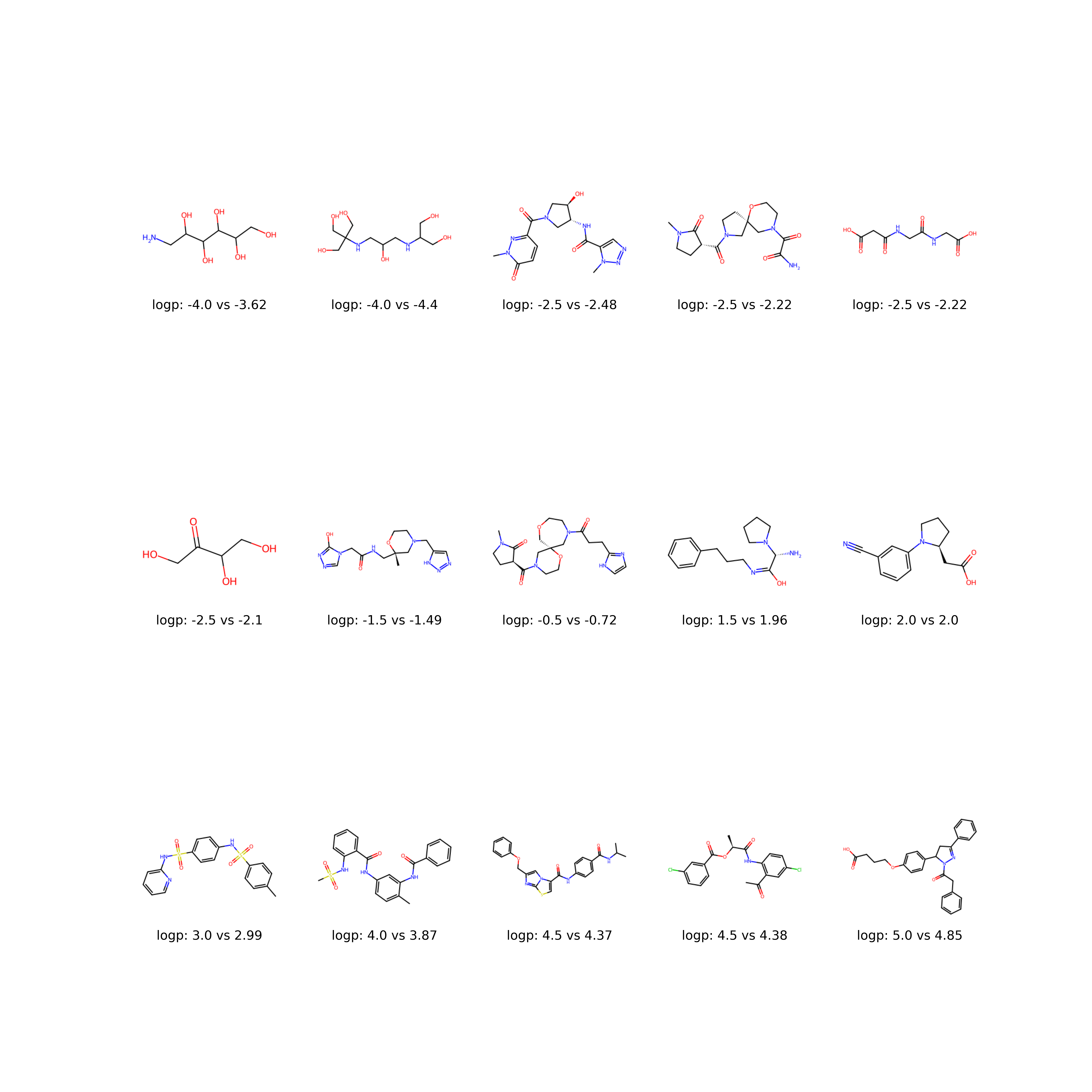}
      \small{\caption{A sample of the generated molecular with logP as conditioning.} \label{fig:logp_sample_viz}}
\end{figure}

\begin{figure} 
    \centering
      \includegraphics[width=1.0\textwidth,height=1.0\textwidth]{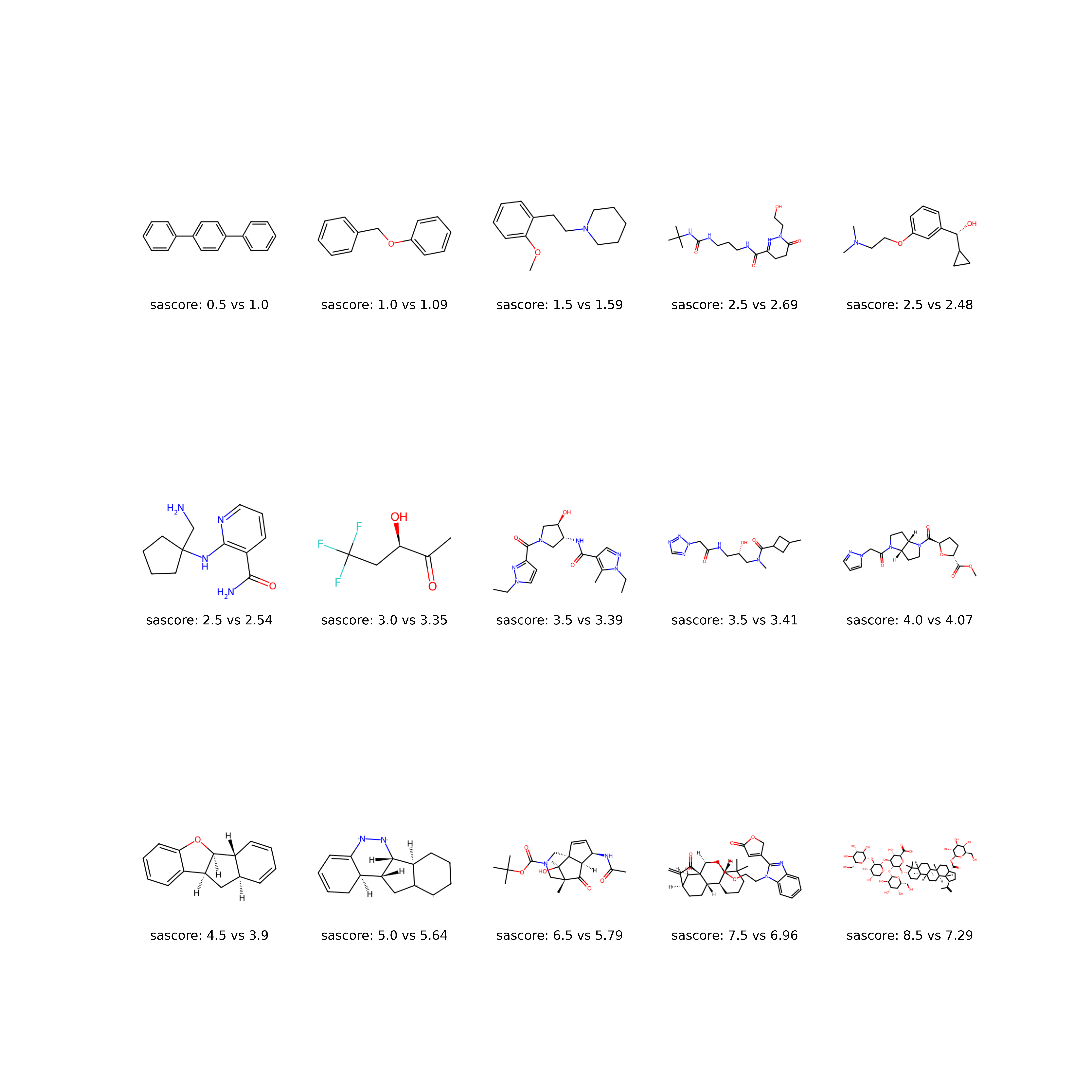}
      \small{\caption{A sample of the generated molecular with SAScore as conditioning.} \label{fig:sas_sample_viz}}
\end{figure}

\begin{figure}
    \centering
      \includegraphics[width=1.0\textwidth,height=1.0\textwidth]{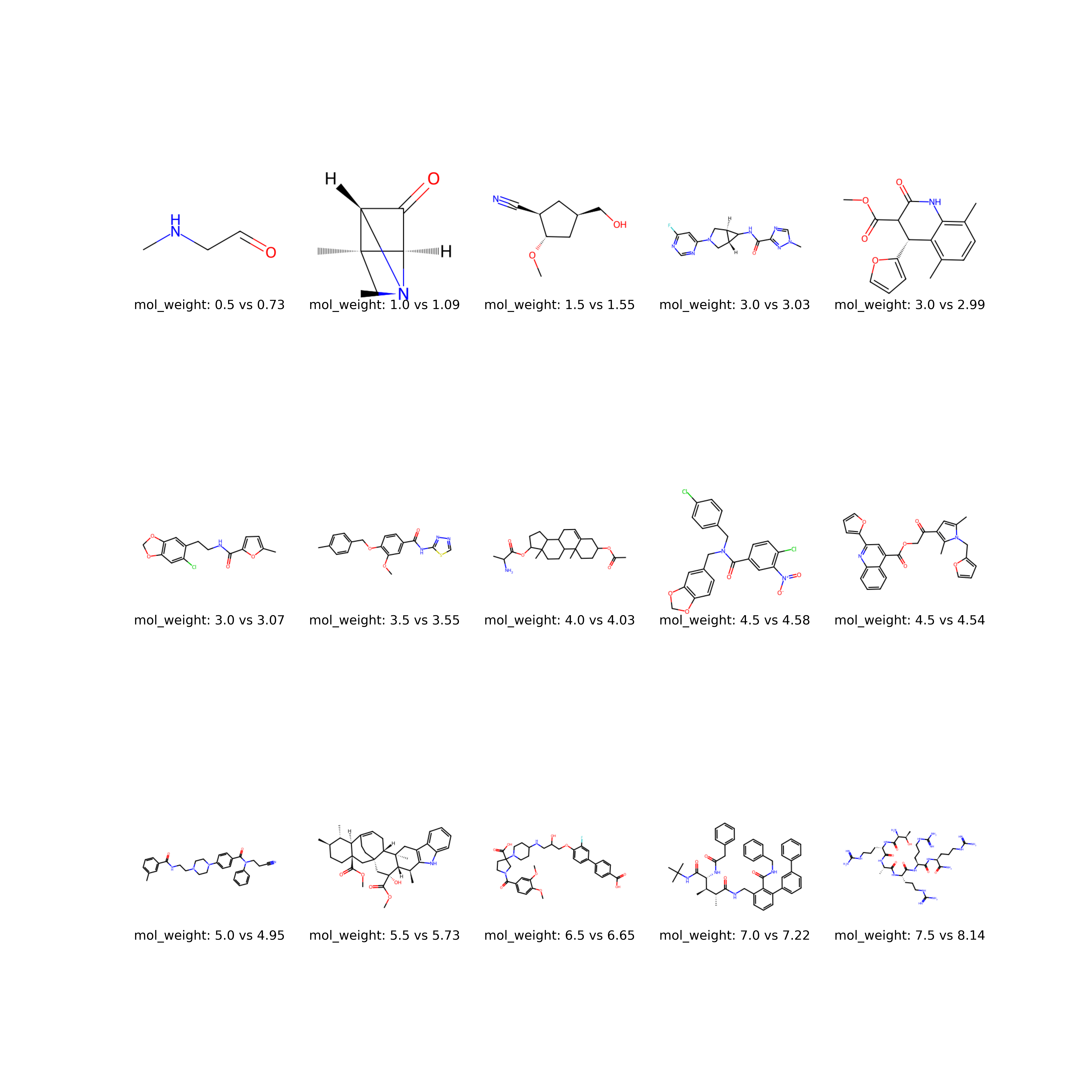}
      \small{\caption{A sample of the generated molecular with the molecular weight as conditioning.} \label{fig:weight_sample_viz}}
\end{figure}

In this section are a sample of the generated molecules for each property visualized.
In Figure \ref{fig:logp_sample_viz} showcases examples that are generated with logP as a property from negative to positive values. Furthermore, the Figure \ref{fig:sas_sample_viz} show the change over different SAScores. Lastly, the Figure \ref{fig:weight_sample_viz} shows how the generated molecules get larger with a rising molecular weight.

\bibliographystyle{unsrt}  
\bibliography{references}

\end{document}